\documentclass[letterpaper, 10 pt, conference]{ieeeconf}

\IEEEoverridecommandlockouts

\usepackage{etextools}
\usepackage{amssymb}
\usepackage{float}
\usepackage{makeidx}
\usepackage{amsmath}
\usepackage{bbm}
\usepackage{epsfig}
\usepackage{epsf}
\usepackage{psfrag}
\usepackage{verbatim}
\usepackage{color}
\usepackage{multirow}
\usepackage{tabularx}
\usepackage{booktabs}
\usepackage[tight,footnotesize]{subfigure}
\usepackage{array}
\usepackage{soul}
\usepackage{footnote}
\usepackage{cite}
\usepackage{dblfloatfix}

\usepackage{color, colortbl}
\usepackage[colorlinks,bookmarksnumbered,citecolor=orange,urlcolor=orange]{hyperref}
\usepackage{graphicx}
\graphicspath{{Figures}}
\DeclareGraphicsExtensions{.pdf,.png}
\usepackage{bigstrut}
\usepackage[english]{babel}

\usepackage{csquotes}

\usepackage[T1]{fontenc}
\usepackage{algorithm, setspace}
\usepackage{algpseudocode}
\usepackage{url}
\usepackage{multirow}
\usepackage{float}
\usepackage{xcolor}
\usepackage{hyperref}
 \hypersetup{
     colorlinks=true,
     linkcolor=orange,
     filecolor=orange,
     citecolor=orange,      
     urlcolor=orange,
     }

\newcommand{\p}[1]{\smallskip \noindent \textbf{{#1}.}}
\newcommand{\eq}[1]{Equation~(\ref{eq:#1})}
\newcommand{\fig}[1]{Figure~\ref{fig:#1}}

\usepackage{balance}

\title{\LARGE

Learning Latent Actions without Human Demonstrations

}

\author{Shaunak A. Mehta*, Sagar Parekh*, and Dylan P. Losey
\thanks{*Shaunak Mehta and Sagar Parekh contributed equally to this work.}
\thanks{The authors are members of the Collaborative Robotics Lab (\href{https://collab.me.vt.edu/}{Collab}), Dept. of Mechanical Engineering, Virginia Tech, Blacksburg, VA 24061.
\newline
{e-mail: \texttt{\{mehtashaunak, sagarp, losey\}@vt.edu}}}
}

\begin{document}
\maketitle

\begin{abstract}

We can make it easier for disabled users to control assistive robots by mapping the user's low-dimensional joystick inputs to high-dimensional, complex actions. Prior works learn these mappings from \textit{human demonstrations}: a non-disabled human either teleoperates or kinesthetically guides the robot arm through a variety of motions, and the robot learns to reproduce the demonstrated behaviors. But this framework is often impractical --- disabled users will not always have access to external demonstrations! Here we instead learn diverse teleoperation mappings \textit{without} either human demonstrations or pre-defined tasks. Under our unsupervised approach the robot first optimizes for object state entropy: i.e., the robot autonomously learns to push, pull, open, close, or otherwise change the state of nearby objects. We then embed these diverse, object-oriented behaviors into a latent space for real-time control: now pressing the joystick causes the robot to perform dexterous motions like pushing or opening. We experimentally show that --- with a best-case human operator --- our unsupervised approach actually \textit{outperforms} the teleoperation mappings learned from human demonstrations, particularly if those demonstrations are noisy or imperfect. But our user study results were less clear-cut: although our approach enabled participants to complete tasks more quickly and with fewer changes of direction, users were confused when the unsupervised robot learned unexpected behaviors. See videos of the user study here: \url{https://youtu.be/BkqHQjsUKDg}

\end{abstract}



\section{Introduction}

\begin{figure}[t]
	\begin{center}
		\includegraphics[width=0.7\columnwidth]{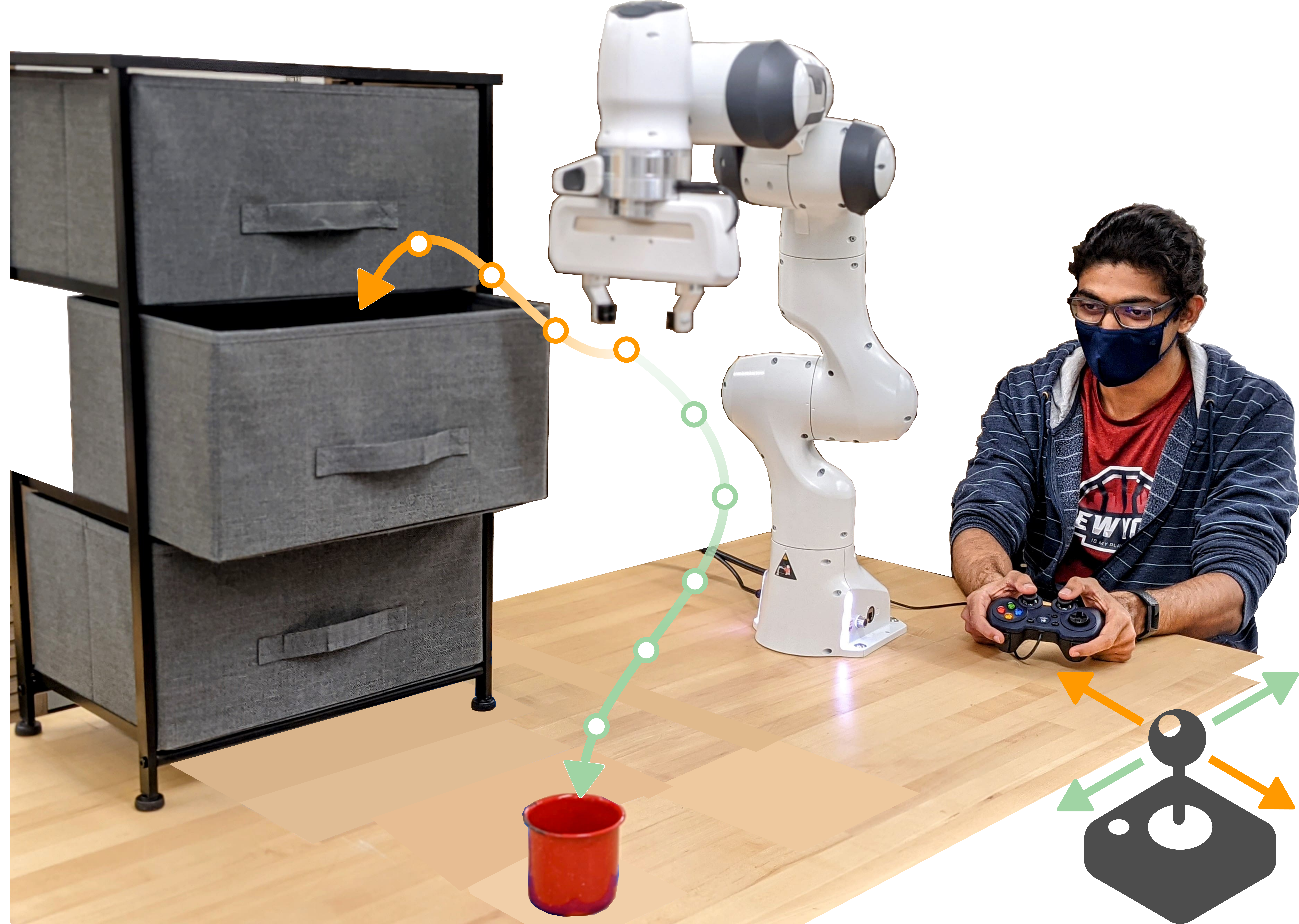}
		\vspace{-0.5em}
		\caption{Human teleoperating an assistive robot arm. We hypothesize that assistive robots can learn useful teleoperation mappings without human supervision by optimizing for high-entropy and object-related behaviors, and then embedding those diverse actions into a user-controlled latent space. Here the robot autonomously learns to map the user's joystick inputs to dexterous, coordinated motions that reach for the cup or open the drawer.}
		\label{fig:front}
	\end{center}
 	\vspace{-2.0em}
\end{figure}

Wheelchair-mounted robot arms have the potential to improve the lives of over one million American adults living with physical disabilities \cite{taylor2018americans}. Imagine teleoperating a wheelchair-mounted robot arm to interact with the environment shown in \fig{front}. You control the motion of the robot arm using a joystick \cite{kinova}, and you have in mind some task that you want to perform. Looking again at \fig{front}, we recognize that there are two likely tasks: reaching for the cup or interacting with the drawer. If the assistive robot also recognizes what tasks are possible within this environment, then it can help you to coordinate the arm's motion. More specifically, the robot can directly map your joystick inputs to complex, task-related behaviors: i.e., pressing down on the joystick causes the robot arm to reach for the cup, and pressing right causes the robot to open the drawer.

Recent research enables assistive robots to learn teleoperation mappings from low-dimensional joystick inputs to high-dimensional robot actions \cite{losey2021learning, karamcheti2021learning}. Within this prior work a non-disabled person \textit{demonstrates} the possible tasks to the robot --- i.e., a caregiver kinesthetically guides the robot through the process of reaching for the cup or opening the drawer. After the non-disabled person provides a variety of different demonstrations, the robot embeds the demonstrated behavior into low-dimensional \textit{latent actions}, which the disabled person then uses to teleoperate the assistive arm. Returning to our example, the user's joystick inputs now map to high-dimensional reaching or opening motions.

This approach to learning latent actions makes sense when a caregiver is available to provide the initial demonstrations. But this is not always possible --- indeed, a key motivation for assistive robots is reducing the user's dependence on external caregivers. In this paper we therefore propose to learn latent actions \textit{without any human supervision}. Our insight is that --- even if the robot does not know what tasks the human might want to perform --- the robot can learn meaningful and diverse latent actions by realising that:
\begin{center}\vspace{-0.4em}
\textit{Humans often use assistive robot to interact with and change the state of objects in the environment.}\vspace{-0.4em}
\end{center}
We apply this insight to train the assistive robot arm to learn fully-autonomous policies that have diverse effects on the world (i.e., we train the robot to maximize object state entropy over repeated interactions). Looking at the environment in \fig{front}, this approach causes the robot to learn behaviors like picking up the cup, moving the cup, opening the drawer, and closing the drawer. We next rollout these unsupervised, diverse behaviors to generate the demonstrations for learned latent actions. Our hypothesis is that --- by learning latent actions which alter the environment state in object-oriented, task-agnostic ways --- we will autonomously acquire a useful and assistive teleoperation mapping.

Overall, we make the following contributions:

\p{Formalize Unsupervised Pre-Training for Latent Actions} Our two-step approach trains the robot to maximize object entropy, and then uses an autoencoder to embed these diverse behaviors into a latent space the human can control. We emphasize the fundamental assumptions behind this approach.

\p{Compare with Human-Provided Demonstrations} We collect kinesthetic and teleoperated demonstrations from participants. We show that the latent actions learned with our unsupervised approach result in more successful task completion than latent actions trained on human demonstrations, particularly if the human demonstrations are noisy.

\p{Apply in New and Unseen Environments} We conduct a user study where non-disabled participants teleoperate the robot arm using our approach and an industry-standard baseline. Participants must combine multiple latent actions and generalize to new object locations. Our results indicate that unsupervised assistance reduces the overall task time.
\section{Related Work}

In this paper we propose to leverage unsupervised pre-training as a way to learn latent actions without requiring the human caregiver or disabled user to provide demonstrations.

\p{Assistive Robots} Wheelchair-mounted robot arms can help disabled adults perform activities of daily living without relying on an external caregiver \cite{mitzner2018closing, argall2018autonomy}. To accomplish these tasks, assistive arms must be high-dimensional and dexterous. But because it is challenging to directly control every individual aspect of the robot's motion \cite{herlant2016assistive}, prior work suggests that disabled adults prefer assistive arms with \textit{partial} or \textit{shared} autonomy \cite{bhattacharjee2020more, gopinath2016human}. Here the human uses an interface (e.g., a joystick) to indicate their desired task, and the robot provides autonomous assistance or guidance to help the human complete that task \cite{reddy2018shared, javdani2018shared,jain2019probabilistic,jonnavittula2021learning}. Consistent with prior work, we also develop a partially autonomous framework that keeps the human in control while autonomously coordinating the motion of the robot arm.

\p{Latent Actions} More specifically, we learn a mapping from low-dimensional human commands (e.g., 2-DoF joystick inputs) to high-dimensional robot actions (e.g., 7-DoF joint velocities). Within the state-of-the-art, a non-disabled caregiver uses kinesthetic demonstrations to show the robot arm how to perform a variety of tasks \cite{losey2021learning, karamcheti2021learning}. The robot then embeds these dexterous, high-dimensional demonstrations into low-DoF \textit{latent actions}, and the disabled user controls the robot with these latent actions. This is analogous to performing principal component analysis on the expert dataset and letting the human control the robot using the first few eigenvectors. Although our work is most closely related to \cite{losey2021learning}, \cite{karamcheti2021learning}, and \cite{lynch2020learning}, we also recognize similarities with reinforcement learning approaches where the robot learns a latent space from expert human demonstrations, and then leverages this latent space to autonomously perform new tasks without a human-in-the-loop \cite{pertsch2020accelerating, singh2020parrot, shankar2020learning, merel2018neural, hausman2018learning, ajay2020opal}. The key difference here is that we will learn latent actions \textit{without expert demonstrations} --- i.e., without requiring the disabled user or human caregiver to first show example motions to the robot.

\p{Unsupervised Pre-Training} Instead of collecting demonstrations from a human, we propose to use unsupervised pre-training. Here the robot arm \textit{learns diverse behaviors} without being given a specific task to accomplish. For example, we can encourage the robot to optimize for policies that behave in unexpected ways (i.e., maximizing prediction error) \cite{pathak2017curiosity}, to learn skills that are very different from one another (i.e., maximizing mutual information) \cite{eysenbach2019diversity, sharma2020dynamics}, or to visit a wide variety of different states (i.e., maximizing state entropy) \cite{hazan2019provably, liu2021behavior}. Works on reinforcement learning have leveraged the policies generated by unsupervised pre-training as priors for downstream tasks \cite{lee2021pebble}. But we explore a fundamentally different setting: we enable the \textit{human} to control the robot by mapping their inputs to the diverse learned behaviors.

\section{Problem Statement}

We consider scenarios where a human is teleoperating their assistive robot arm in household environments. The human controls the robot with a joystick interface, and the robot must interpret the human's low-dimensional commands to perform meaningful actions that assist the human.

\p{Environments} As the wheelchair-mounted robot arm moves around the house the user will inevitably encounter a variety of environments (e.g., eating in the kitchen or working at a desk). We formulate each environment as an undiscounted Markov decision process without rewards: $\mathcal{M} = \langle\mathcal{S}, \mathcal{A}, T\rangle$. Here $s \in \mathcal{S}$ is the system state, $a \in \mathcal{A} \subseteq \mathbb{R}^n$ is the robot action, and $T(s, a)$ captures the dynamics. The action $a$ is high-dimensional: in our experiments $a \in \mathcal{A} \subseteq \mathbb{R}^n$ is the joint velocity of the $n$-DoF robot arm. But the state $s$ is a higher dimension: it includes both the \textit{robot's state} (e.g., its joint position) and the \textit{state of objects in the environment} (e.g., visual observations from an RGB camera). Within our experiments we simplify this by assuming that we have direct access to the object states --- i.e., we know their position and orientation\footnote{This simplification matches recent work on latent actions where the robot leverages a visual object detection model to extract object locations \cite{karamcheti2021learning}.}. Hence, we write the state as $s = (s_{\mathcal{R}}, s_{\mathcal{O}})$, where $s_{\mathcal{R}} \in \mathbb{R}^n$ is the robot's joint position, $o_i$ is the pose of the $i$-th object, and $s_{\mathcal{O}} = (o_1, o_2, \ldots, o_K) \in \mathbb{R}^m$ is a vector that includes the pose of each object in the environment.

Recall that the human and robot will interact in several different environments. More formally, let $p(\mathcal{M})$ denote a distribution over environments, so that as the human moves around the house they sample environments $\mathcal{M} \sim p(\mathcal{M})$. We note that $\mathcal{M}$ \textit{does not include a reward function} because we do not know what task (or tasks) the human will want to perform within each environment.

\begin{figure*}[t]
	\begin{center}
		\includegraphics[width=2.0\columnwidth]{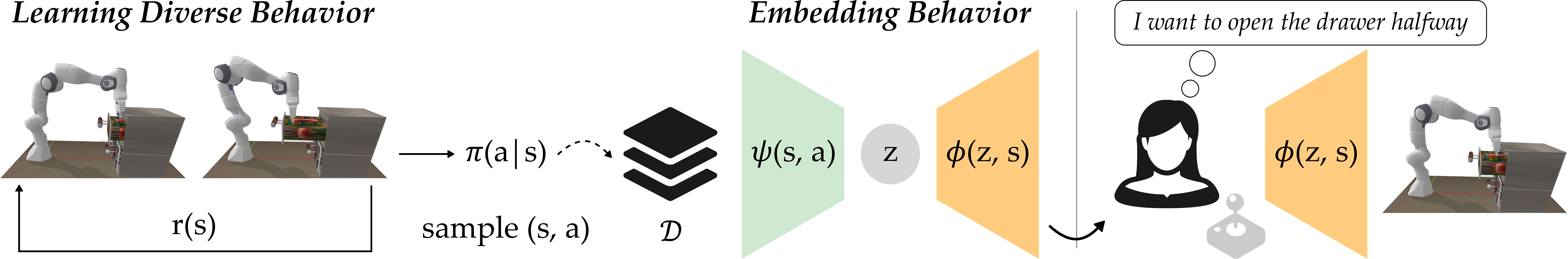}
		\vspace{-0.5em}
		\caption{Our proposed approach for learning assistive teleoperation mappings without human supervision. (Left) Within a simulated environment the robot arm learns autonomous behaviors that maximize object entropy --- e.g., the robot learns to open the drawer different amounts. (Middle) We then sample trajectories from the trained robot, and aggregate state-action pairs across these trajectories to form the dataset $\mathcal{D}$. The robot embeds these diverse, object related motions into latent actions by leveraging a state-conditioned autoencoder. (Right) We apply the learned decoder with a human-in-the-loop. The decoder maps the human's joystick inputs to coordinated, object-related actions --- e.g., the human controls exactly how far the robot opens the drawer.}
		\label{fig:method}
	\end{center}
	\vspace{-2em}
\end{figure*}

\p{Teleoperation} Returning to our example environment from \fig{front}, the robot does not know whether the human wants to open the drawer, pick up the cup, or accomplish some other unexpected task. To convey their intent the disabled user teleoperates the robot arm by applying \textit{joystick inputs} $u \in \mathcal{U} \subseteq \mathbb{R}^d$. We recognize that there are many teleoperation devices --- such as sip-and-puff tools \cite{argall2018autonomy} or brain-computer interfaces \cite{muelling2015autonomy} --- but here we focus on joysticks since they are the most prevalent input modality for today's wheelchair-mounted robot arms \cite{kinova}. Importantly, the dimension of the user's joystick is lower than the dimension of the robot's action space ($d < n$). Hence, we need a mapping that converts the human's \textit{low-dimensional} joystick inputs into \textit{high-dimensional} robot actions.

\p{Latent Actions} Prior work on shared autonomy assumes a pre-defined teleoperation mapping with multiple modes \cite{newman2018harmonic, herlant2016assistive}. Consider using a $2$-DoF joystick to control a robot arm: your joystick moves the robot's end-effector along the $x$-$y$ axes in one mode, in $z$-$roll$ axes in another mode, and so on. By contrast, we seek to \textit{learn a projection function} that maps low-dimensional user inputs into high-dimensional, task-related robot motions. Inspired by recent research on latent actions \cite{losey2021learning}, we formulate this mapping as a \textit{decoder}:
\begin{equation} \label{eq:decoder}
    a = \phi(z, s)
\end{equation}
Here $z \in \mathcal{Z} \subseteq \mathbb{R}^d$ is a low-dimensional latent action that the human can directly input using their $d$-dimensional joystick. The decoder $\phi$ takes in this input as well as the system state $s$, and outputs a high-dimensional robot action $a$. Our objective is to \textit{learn} a decoder $\phi$ that enables the robot to perform a variety of useful actions across environments $M \sim p(M)$ so that the human can seamlessly control the robot. But our challenge is learning this decoder\textit{ without supervision}: how do we identify meaningful latent actions in the absence of human-provided demonstrations or pre-defined tasks?
\section{Unsupervised Latent Actions}

To learn an assistive teleoperation mapping we return to our insight: humans often use robots to alter the state of objects around them. Put another way, the human's joystick inputs should cause the robot to perform actions like reaching for, picking up, moving, opening, or rotating \textit{nearby objects}. Of course, not all of these actions apply to every object --- an opening motion might change the state of a drawer, but will have no effect on a cup. We therefore propose an approach where the robot samples environments $\mathcal{M} \sim p(\mathcal{M})$, and \textit{autonomously learns} diverse actions for the specific objects in those environments (see Section~\ref{sec:RL}). We then embed these diverse behaviors into a low-dimensional latent action space (see Section~\ref{sec:IL}). Overall, our unsupervised approach outlined in \fig{method} learns latent actions that alter nearby objects (e.g., pressing right to increasingly open the drawer), enabling the human to teleoperate the robot across a variety of object-related motions without ever requiring pre-specified tasks or human demonstrations. We emphasize that both Sections~\ref{sec:RL} and \ref{sec:IL} occur without a human-in-the-loop: our final output is the learned decoder in \eq{decoder}, which is then leveraged by the human to control the robot.

\subsection{Learning Diverse Behavior} \label{sec:RL}

\p{Intrinsic Reward} Given one or more environments $\mathcal{M} \sim p(\mathcal{M})$, we will leverage unsupervised pre-training to identify diverse behaviors. Recall that $\mathcal{M}$ does not include a reward function since we do not know what tasks the human has in mind. Instead, we here specify an \textit{intrinsic} reward function that encourages the robot to do two things: (a) maximize the \textit{object state entropy} and (b) minimize the \textit{distance to objects}. Recall that $s_{\mathcal{O}}$ contains the state of each object in the environment. We want to alter those objects in diverse ways \cite{pathak2017curiosity, hazan2019provably, eysenbach2019diversity, sharma2020dynamics}, i.e., we want to maximize the entropy over $p(s_{\mathcal{O}})$. But to interact with objects the robot must first reach them: hence, we shape the intrinsic reward by minimizing the distance between the robot's end-effector and the closest object $o \in s_{\mathcal{O}}$. This leads to the reward function:
\begin{equation}
    r(s) = \mathcal{H}(s_{\mathcal{O}}) - \min_{o \in s_{\mathcal{O}}}~{d(s_{\mathcal{R}}, o)}
\end{equation}
where $\mathcal{H}$ is the Shannon entropy and $d(s_{\mathcal{R}}, o)$ is the distance between the robot's end-effector and object $o$. Since computing state entropy is typically intractable, we approximate it using the particle-based estimate from Liu and Abbeel \cite{liu2021behavior}:
\begin{equation} \label{eq:reward}
        r(s) \approx \log{\|s_{\mathcal{O}} - s_{\mathcal{O}}^k\|} - \min_{o \in s_{\mathcal{O}}}~{d(s_{\mathcal{R}}, o)}
\end{equation}
Here $s_{\mathcal{O}}$ is the current state of objects in the environment and $s_{\mathcal{O}}^k$ is the $k$-th nearest neighbor. As the robot interacts with the environment it maintains a replay buffer of recent object states: to compute $s_{\mathcal{O}}^k$ the robot simply searches through this buffer. Intuitively, the first term in \eq{reward} rewards the robot for moving objects into a state $s_{\mathcal{O}}$ that is very different than the object states the robot has recently observed.

\p{Reinforcement Learning} Under \eq{reward} the robot is constantly seeking new and unique object states. Returning to our example, imagine that the robot \textit{closed} the drawer during the previous interaction --- during the next interaction, the robot can increase its reward by \textit{opening} the drawer. We accordingly use reinforcement learning to identify autonomous robot behaviors that maximize the discounted sum of rewards: $\sum_{t=0}^{\infty} \gamma^t  \cdot r(s_t)$. Specifically, we apply \textit{Soft Actor-Critic} (SAC), an off-policy reinforcement learning approach for continuous state and action spaces \cite{haarnoja2018soft}. When using our reward from \eq{reward}, SAC trains the robot to take autonomous actions $a$ that maximize the object state entropy across the replay buffer. The output of the first part of our approach is therefore a \textit{learned robot policy} $\pi(a \mid s)$ that generates diverse, object-related motions.

\subsection{Embedding Diverse Behavior to Latent Actions} \label{sec:IL}

\p{Dataset} In the second part of our approach we leverage the learned behavior from Section~\ref{sec:RL} to train latent actions. We start with policy $\pi(a \mid s)$, the result of unsupervised pre-training across one or more environments $\mathcal{M}\sim p(\mathcal{M})$. We repeatedly rollout this policy to generate robot trajectories $\xi = ((s_0, a_0), (s_1, a_1), \ldots (s_T, a_T))$. Because the robot has been trained to maximize object state entropy, each of these trajectories should have a different effect on objects in the environment: e.g., one trajectory pushes the cup forward, another picks it up, and a third opens the drawer. Finally, we aggregate the state-action pairs across each trajectory to form a cumulative dataset $\mathcal{D} = \{(s_0, a_0), (s_1, a_1), \ldots\}$. Of course, previous research on latent actions also utilizes a dataset --- but the key novelty here is that $\mathcal{D}$ is autonomously generated by the robot, and does not require any human demonstrations.

\p{Embedding} Now that we have a dataset of diverse, object-related actions, we will embed these high-dimensional actions into a low-dimensional latent space. Here we match prior work on latent actions \cite{losey2021learning}, and leverage a conditional autoencoder \cite{doersch2016tutorial}. The \textit{encoder} $\psi : \mathcal{S} \times \mathcal{A} \rightarrow \mathcal{Z}$ embeds the demonstrated behavior into the latent space, and the \textit{decoder} $\phi : \mathcal{Z} \times \mathcal{S} \rightarrow \mathcal{A}$ from \eq{decoder} uses the human's joystick inputs (i.e., latent actions) to reconstruct robot actions. We simultaneously train the encoder and decoder to minimize the action reconstruction loss across the dataset:
\begin{equation} \label{eq:loss}
    \mathcal{L} = \sum_{(s, a) \in \mathcal{D}} \big\| a - \phi\big(\psi(s, a), s\big) \big\|^2
\end{equation}
Finally, we give the trained decoder to the human so that they can teleoperate the robot arm. To understand why this approach works, it is important to remember that the decoder is \textit{conditioned} on state $s$ (which includes both the robot state $s_{\mathcal{R}}$ and the object states $s_{\mathcal{O}}$). Hence, the way that the robot interprets the human's inputs depends on the object locations: e.g., if the drawer is moved to the right, now the robot will reach right (and not forwards) to open this drawer.

\subsection{Assumptions}

Our approach to assistive teleoperation reduces the reliance on external caregivers. However, there is no free lunch --- and here we want to emphasize the \textit{two assumptions} that enable us to learn latent actions without human supervision.

\p{Access to (Simulated) Environments} First, we assume that we know some of the environments the human will interact with \textit{a priori}, i.e., we can draw samples $\mathcal{M} \sim p(\mathcal{M})$. During implementation we form simulated versions of these environments to run the unsupervised pre-training from Section~\ref{sec:RL}. We can mitigate this assumption by leveraging large-scale interactive simulations of home environments \cite{shen2020igibson}.

\p{Access to Object State} Second, we assume that the robot can measure the state of objects, i.e., the robot observes $s_{\mathcal{O}}$. Prior work on latent actions uses pre-trained object detection models \cite{redmon2016you} to obtain the object state from RGB images \cite{karamcheti2021learning}. Object detection and classification is also suitable for our proposed approach; however, we recognize that this may fail if the robot's view of the objects is obstructed.
\section{Comparison to Human Demonstrations}

\begin{figure*}[t]
	\begin{center}
		\includegraphics[width=2.0\columnwidth]{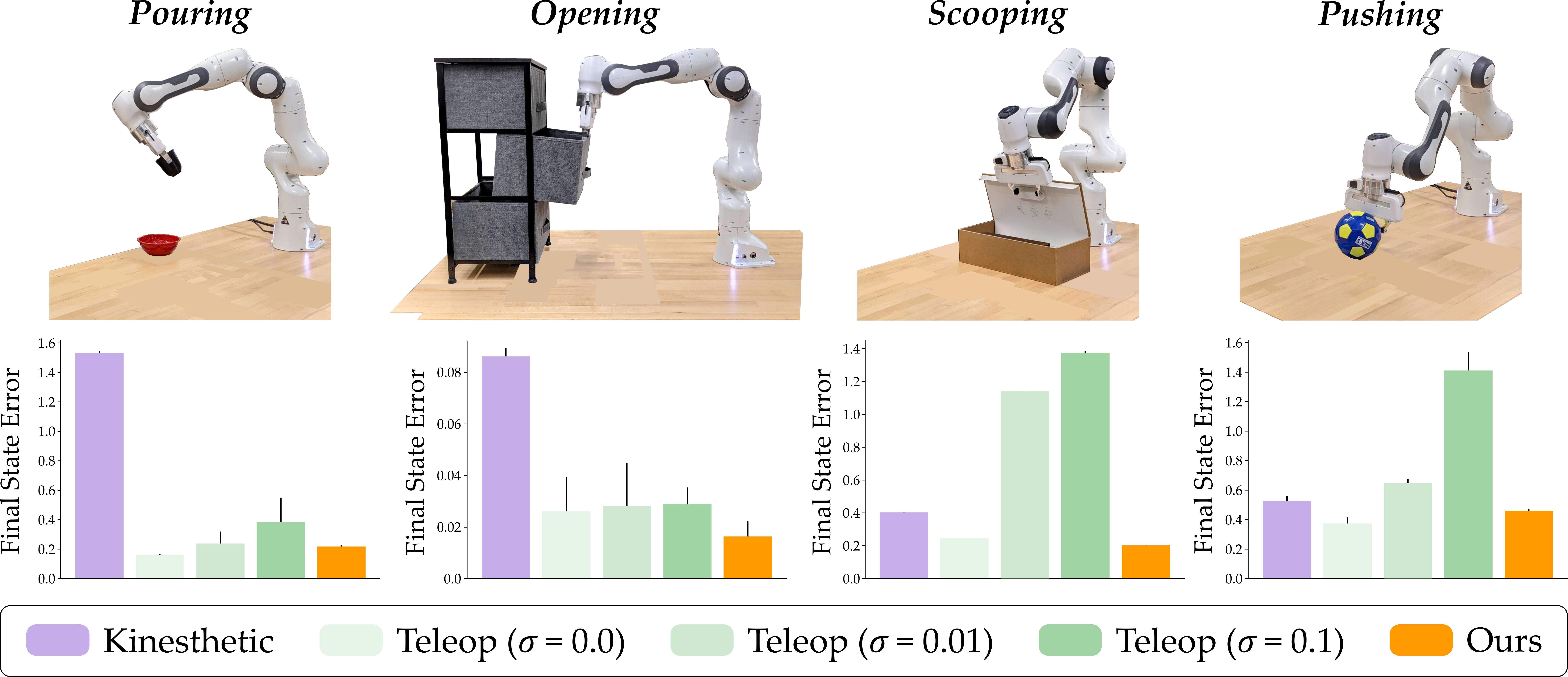}
		\vspace{-0.5em}
		\caption{Simulated human controlling the Panda robot arm with learned latent actions. We compare latent actions trained on human-provided kinesthetic demonstrations (\textbf{Kinesthetic}) and teleoperated demonstrations (\textbf{Teleop}) to our unsupervised approach (\textbf{Ours}). We also consider \textbf{Teleop} as the participants' demonstrations become increasingly noisy and imperfect ($\sigma = 0.01$, $\sigma = 0.1$). \textbf{Ours} outperforms \textbf{Kinesthetic} and noisy versions of \textbf{Teleop}, and is slightly worse than \textbf{Teleop} ($\sigma = 0.0$). Our method enables the robot to match prior latent action approaches without requiring any human demonstrations.}
		\label{fig:sim1}
	\end{center}
	\vspace{-2.0em}
\end{figure*}

We have formalized an approach to learning latent actions without human-provided demonstrations. However, it is not yet clear how these \textit{unsupervised} latent actions compare to \textit{supervised} latent actions (i.e., latent actions trained on human demonstrations). Here we collect offline demonstrations from study participants using a Panda robot arm. We consider both \textit{kinesthetic} demonstrations --- where participants physically guide the robot through the tasks --- and \textit{teleoperated} demonstrations --- where participants use a joystick to directly control the robot's end-effector. We then train latent actions on the human-provided datasets, and compare the resulting teleoperation mappings to our unsupervised approach. 

\p{Human Demonstrations} We recruited $7$ non-disabled participants (ages $24 \pm 3.5$) to provide both \textbf{Kinesthetic} and teleoperated (\textbf{Teleop}) demonstrations. In \textbf{Kinesthetic} participants physically guided the robot arm: these demonstrations are suitable for a non-disabled caregiver. By contrast, in \textbf{Teleop} participants controlled the robot's end-effector with a joystick: in practice, these demonstrations could be provided directly by the disabled user.

To understand how human mistakes affect the latent actions learned from these datasets, we also considered \textit{noisy versions} of \textbf{Teleop}. Here we added zero-mean Gaussian noise to the participants' demonstrated actions. We tested three levels of standard deviation, from $\sigma = 0.0$ (i.e., no noise), to $\sigma = 0.01$ and $\sigma = 0.1$.

\p{Decoders} After collecting human demonstrations, we used these demonstrations to train decoders (i.e, latent actions). More specifically, we formed separate datasets from \textbf{Kinesthetic} and \textbf{Teleop}, and applied the approach from Section~\ref{sec:IL} to embed these datasets into latent actions. We compared the resulting decoders to the output of our proposed approach (\textbf{Ours}). To avoid biasing the results towards our method, we made sure to collect more state-action pairs from human demonstrations than from unsupervised pre-training. On average, $|\mathcal{D}| = 23.5$k for \textbf{Kinesthetic}, $|\mathcal{D}| = 16.3$k for \textbf{Teleop}, and $|\mathcal{D}| = 14$k for \textbf{Ours}.

\p{Simulated Controller} We controlled the robot using a simulated human to standardize the \textit{best-case} performance of each approach. This simulated human was given a goal state $s^*$, and selected greedily optimal latent actions $z$ to move the Panda robot arm towards that goal state:
\begin{equation} \label{eq:simhuman}
    z = \text{arg}\min_{z \in \mathcal{Z}} \big\| s^* - T\big(s, \phi(z, s)\big) \big\|^2
\end{equation}
Recall that $T(s, a)$ in \eq{simhuman} is the system dynamics. Although we used a simulated human to control the robot, \textit{these comparisons were all performed on a real robot arm}.

\p{Environments} We started with the settings in \fig{sim1}.

\begin{itemize}
    \item \textit{Pouring}: $s_{\mathcal{O}}$ contains the pose of the cup and bowl, and the human's goal is to pour the cup above the bowl.
    \item \textit{Opening}: $s_{\mathcal{O}}$ contains the position of the drawers, and the human's goal is to open the drawer.
    \item \textit{Scooping}: $s_{\mathcal{O}}$ contains the position and angle of the lid, and the human's goal is to open the lid.
    \item \textit{Pushing}: $s_{\mathcal{O}}$ contains the position of the ball, and the human's goal is to push the ball to three locations.
\end{itemize}    
For \textit{Pouring}, \textit{Opening}, and \textit{Scooping} we embeded the robot's actions into a $1$-DoF latent space (i.e., the simulated human can only press right or left on the joystick), and for \textit{Pushing} we leveraged a $2$-DoF latent space.

\p{Results} Within each environment the simulated human attempted to reach a goal state. We measured the error between this goal state and the closest state that the robot actually reached: our results are displayed in \fig{sim1}.

Across all four environments the latent actions learned from \textbf{Kinethetic} demonstrations were less accurate than \textbf{Ours}\footnote{Different participants provided different kinesthetic demonstrations for the same task (e.g., moving all the joints or just the last joint when scooping). This demonstration variability caused the \textbf{Kinesthetic} to perform worse than \textbf{Teleop}, where the constraint of teleoperating the robot's end-effector forced different participants to provide similar demonstrations.}. Without any noise (i.e., $\sigma = 0$), the robots using \textbf{Teleop} had the lowest final state error ($p < .05$). But as the amount of noise increased \textbf{Ours} again outperformed the \textbf{Teleop} baseline. Viewed together, these results suggest that --- \textit{without any human demonstrations} --- our approach learns teleoperation mappings that are just as effective as latent actions trained on human-provided demonstrations.

\begin{figure}[t]
    \vspace{1em}
	\begin{center}
		\includegraphics[width=0.9\columnwidth]{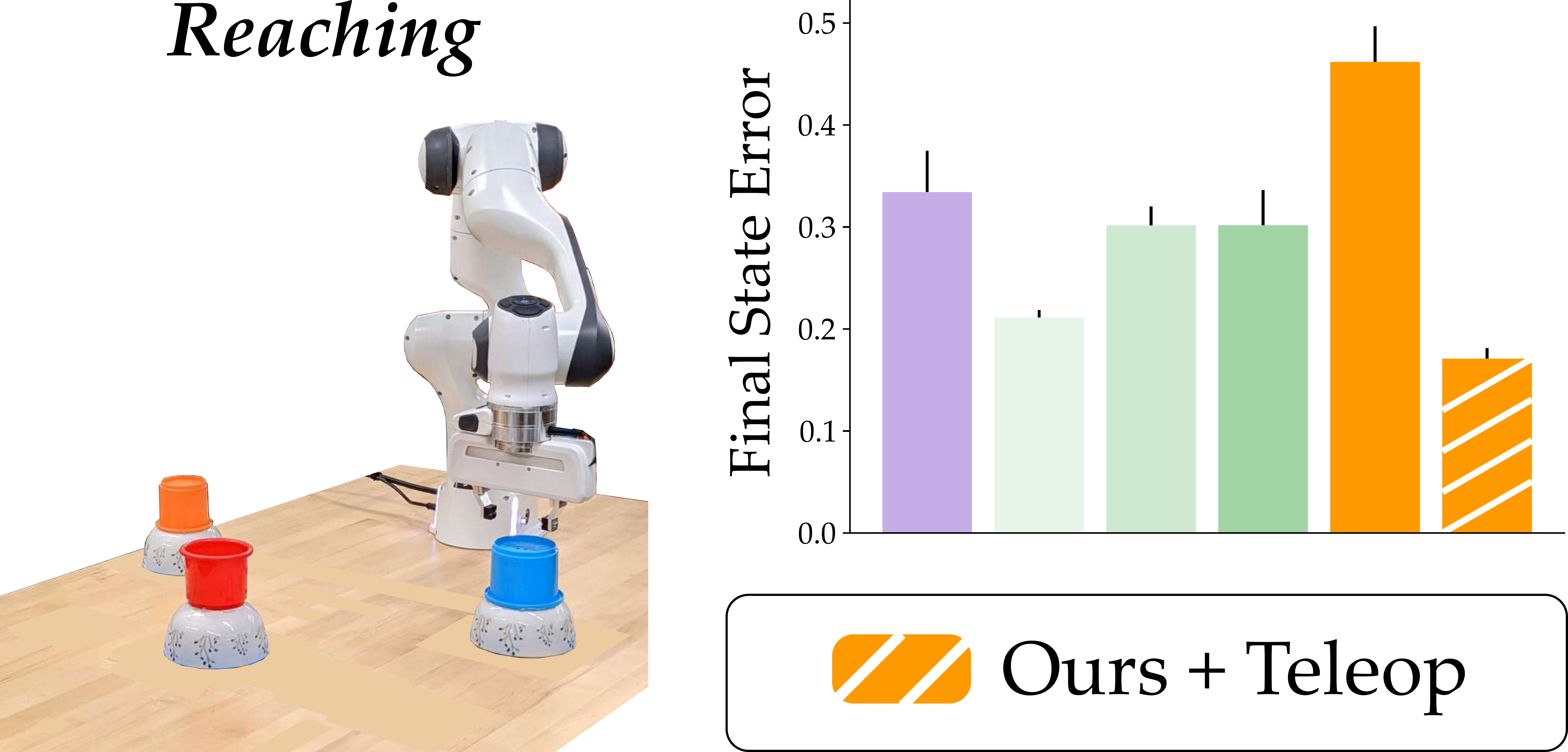}
		\vspace{-0.5em}
		\caption{Follow-up experiment where we initialize our approach with human demonstrations (\textbf{Ours + Teleop}). These demonstrations help our approach avoid local minima (i.e., only reaching two cups), and our approach improves the demonstrations by optimizing for other high-entropy behaviors.}
		\label{fig:sim2}
	\end{center}
	\vspace{-2em}
\end{figure}

\begin{figure*}[t]
	\begin{center}
		\includegraphics[width=2\columnwidth]{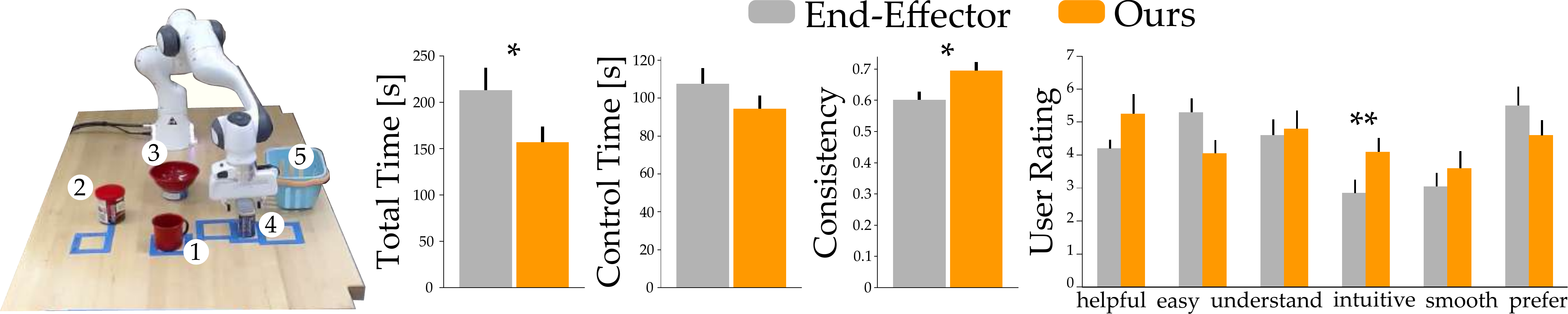}
		\vspace{-0.5em}
		\caption{Task and results from our user study. Participants teleoperated the robot arm to remove clutter (objects 1 and 2), pull the bowl closer to their person (3), pick up the container (4) and pour it into the bowl, and finally drop the container in the bin (5). We compare our unsupervised approach (where participants interact with one joystick) to direct end-effector control (where participants can use two joysticks and toggle between two modes). Participants completed the task more quickly with our approach, but their subjective responses were mixed. Here $*$ denotes $p<.05$ and $**$ denotes $p<.07$.}
		\label{fig:results}
	\end{center}
	\vspace{-2em}
\end{figure*}

\p{Follow-up} Can we use human demonstrations to improve unsupervised latent actions? The robot will inevitably reach scenarios where human demonstrations are available. Rather than discarding this data, we propose to \textit{combine} both human demonstrations and our unsupervised approach. Recall that the robot maintains a replay buffer when learning diverse behaviors in Section~\ref{sec:RL}: here we initialize the robot's replay buffer with human-provided demonstrations. This seeds the robot's search with the diverse behaviors that the human has already shown, enabling the robot to learn additional motions that build on the demonstrated behaviors.

We conducted a follow-up experiment in the \textit{Reaching} environment (see \fig{sim2}). As before, a simulated human controlled the robot: we embedded the robot's actions into a $1$-DoF latent space, and the simulated human leveraged this latent space to reach for three cups on the table. But unlike the previous environments, here \textbf{Ours} performed the worst. This is because our unsupervised pre-training approach got stuck in a local minima (and only learned to reach for two of the cups). Initializing with teleoperated demonstrations where the human guided the robot to all three cups solved this problem: \textbf{Ours + Teleop} outperformed the baselines, including the original \textbf{Teleop} demonstrations. 
\section{User Study}

During our comparison experiments the robot was only interacting with a single object (e.g., a cup, drawer, or ball). To evaluate our approach in more realistic scenarios, we conducted a user study with non-disabled participants: here the robot had to generalize to \textit{multiple objects} in \textit{previously unseen} locations (see \fig{results} and supplementary \href{https://youtu.be/BkqHQjsUKDg}{video}).

\p{Experimental Setup} Participants teleoperated the 7-DoF robot arm (\href{https://www.franka.de/robot-system}{Panda, FrankaEmika}) using one $2$-DoF joystick (Logitech F310 Gamepad). Participants had to (a) move clutter out of the way, (b) pull the bowl to their side of the table, (c) pick up a container, (d) pour the container above the bowl and (e) drop the empty container into a bin. We rearranged the locations of the objects between trials.

\p{Independent variables} We compared our unsupervised latent actions (\textbf{Ours}) to an industry-standard baseline (\textbf{End-Effector}) \cite{kinova}. Within \textbf{End-Effector} participants directly controlled the velocity of the robot's end-effector. They pressed a button to toggle between two different modes: one mode controlled the robot's linear velocity, and the other controlled the robot's angular velocity \cite{herlant2016assistive, newman2018harmonic}. By contrast, with \textbf{Ours} the robot mapped the human's joystick inputs to learned joint velocities. Similar to \textbf{End-Effector}, the user could press a button to toggle between two different latent action spaces. We trained \textbf{Ours} offline using a simulated version of the environment (i.e., without human demonstrations).

\p{Dependent Variables} We measured the \textit{Total Time} it took for each participant to complete the task and the amount of time users spent interacting with the joystick (\textit{Control Time}). We also recorded the joystick input $u$ at each timestep, and \textit{Consistency} is the average alignment between these inputs: $(u^{t-1} \cdot u^t) / (\|u^{t-1}\| \|u^{t}\|)$. Lower \textit{Consistency} indicates that the human often changed directions when using the joystick. Finally, we asked subjects to complete a 7-point Likert scale survey after finishing each condition. Our survey questions were arranged into six multi-item scales: how \textit{helpful} the robot was, how \textit{easy} it was to control the robot, whether the robot seemed to \textit{understand} the user's goal, how \textit{intuitive} the control interface was, whether the robot moved \textit{smoothly}, and if they \textit{preferred} using that condition.
 
\p{Participants and Procedure} We recruited $10$ participants from the Virginia Tech community ($5$ female, average age $24.2$ ± $2.9$ years). Each participant provided informed written consent prior to the experiment. We utilized a counterbalanced, within-subject design: each participant completed the task twice with \textbf{Ours} and twice with \textbf{End-Effector}. Between trials we changed the locations of the objects (these changes were identical across both conditions). Half of the participants started with \textbf{Ours}. Prior to each condition, participants were given up to $5$ minutes of practice to familiarize themselves with the teleoperation mapping.

\p{Hypotheses} We had two hypotheses:\\
\textbf{H1.} \textit{Non-disabled users will complete the task more quickly when using unsupervised latent actions.}\\
\textbf{H2.} \textit{Non-disabled users will perceive robots that leverage unsupervised latent actions as better partners.}

\p{Results} The results of our user study are shown in \fig{results}. We found support for \textbf{H1}: participants completed the task in less total time with \textbf{Ours} ($F(1,19)=12.2$, $p<0.01$) and participants maintained more consistent joystick inputs with fewer changes of direction ($F(1,19)=14.7$, $p<0.01$).

For our survey results we first confirmed the reliability of the six scales, and then grouped these scales into a combined score. The resulting comparisons were inconclusive. None of differences were statistically significant, although we found that participants thought \textbf{Ours} was marginally more \textit{intuitive} than the baseline ($p < .07$). The other scales favor \textbf{Ours}, with the exceptions of \textit{easy} and \textit{prefer}. Overall, we were unable to make any conclusions about \textbf{H2}.

\p{Limitations} During the user study our unsupervised mapping occasionally performed actions that participants did not want. For example, \textbf{Ours} learned to push objects off the table. Although this behavior matches our intrinsic objective --- i.e., it greatly changes object state --- it misses out on the internal priors or \textit{affordances} that humans have over these objects. We believe that the unexpected, additional behaviors learned with \textbf{Ours} confused the people using this method (and contributed to mixed user responses). Moving forward
we plan to encode affordances into the learned latent actions; e.g., the robot should know never to knock over a glass vase.
\section{Conclusion}

\noindent We enabled assistive robot arms to learn teleoperation mappings without human demonstrations. Under our two-step approach the robot first leverages unsupervised pre-training to optimize for diverse, object-oriented behaviors, and then embeds those behaviors into a latent space for human control. We experimentally found that the resulting decoder is on par with mappings learned from human demonstrations. Our user study results show that people can efficiently leverage this unsupervised approach in settings with multiple objects.


\newpage
\balance
\bibliographystyle{IEEEtran}
\bibliography{IEEEabrv,bibtex}

\begin{thebibliography}{10}
\providecommand{\url}[1]{#1}
\csname url@rmstyle\endcsname
\providecommand{\newblock}{\relax}
\providecommand{\bibinfo}[2]{#2}
\providecommand\BIBentrySTDinterwordspacing{\spaceskip=0pt\relax}
\providecommand\BIBentryALTinterwordstretchfactor{4}
\providecommand\BIBentryALTinterwordspacing{\spaceskip=\fontdimen2\font plus
\BIBentryALTinterwordstretchfactor\fontdimen3\font minus
  \fontdimen4\font\relax}
\providecommand\BIBforeignlanguage[2]{{%
\expandafter\ifx\csname l@#1\endcsname\relax
\typeout{** WARNING: IEEEtran.bst: No hyphenation pattern has been}%
\typeout{** loaded for the language `#1'. Using the pattern for}%
\typeout{** the default language instead.}%
\else
\language=\csname l@#1\endcsname
\fi
#2}}

\bibitem{taylor2018americans}
D.~M. Taylor, ``Americans with disabilities: 2014,'' \emph{US Census Bureau},
  pp. 1--32, 2018.

\bibitem{kinova}
\emph{\href{https://www.kinovarobotics.com/sites/default/files/ULWS-RA-JAC-UG-INT-EN\%20201804-1.0\%20\%28KINOVA\%E2\%84\%A2\%20Ultra\%20lightweight\%20robotic\%20arm\%20user\%20guide\%29_0.pdf}{KINOVA
  Ultra Lightweight Robotic Arm User Guide}}, 2018 (accessed September 1,
  2021).

\bibitem{losey2021learning}
D.~P. Losey, H.~J. Jeon, M.~Li, K.~Srinivasan, A.~Mandlekar, A.~Garg, J.~Bohg,
  and D.~Sadigh, ``Learning latent actions to control assistive robots,''
  \emph{Autonomous Robots}, pp. 1--33, 2021.

\bibitem{karamcheti2021learning}
S.~Karamcheti, A.~J. Zhai, D.~P. Losey, and D.~Sadigh, ``Learning visually
  guided latent actions for assistive teleoperation,'' in \emph{Learning for
  Dynamics and Control}, 2021, pp. 1230--1241.

\bibitem{mitzner2018closing}
T.~L. Mitzner, J.~A. Sanford, and W.~A. Rogers, ``Closing the capacity-ability
  gap: {U}sing technology to support aging with disability,'' \emph{Innovation
  in Aging}, vol.~2, no.~1, 2018.

\bibitem{argall2018autonomy}
B.~D. Argall, ``Autonomy in rehabilitation robotics: {A}n intersection,''
  \emph{Annual Review of Control, Robotics, and Autonomous Systems}, vol.~1,
  pp. 441--463, 2018.

\bibitem{herlant2016assistive}
L.~V. Herlant, R.~M. Holladay, and S.~S. Srinivasa, ``Assistive teleoperation
  of robot arms via automatic time-optimal mode switching,'' in \emph{ACM/IEEE
  International Conference on Human-Robot Interaction}, 2016, pp. 35--42.

\bibitem{bhattacharjee2020more}
T.~Bhattacharjee, E.~K. Gordon, R.~Scalise, M.~E. Cabrera, A.~Caspi, M.~Cakmak,
  and S.~S. Srinivasa, ``Is more autonomy always better? {E}xploring
  preferences of users with mobility impairments in robot-assisted feeding,''
  in \emph{ACM/IEEE International Conference on Human-Robot Interaction}, 2020,
  pp. 181--190.

\bibitem{gopinath2016human}
D.~Gopinath, S.~Jain, and B.~D. Argall, ``Human-in-the-loop optimization of
  shared autonomy in assistive robotics,'' \emph{IEEE Robotics and Automation
  Letters}, vol.~2, no.~1, pp. 247--254, 2016.

\bibitem{reddy2018shared}
S.~Reddy, A.~D. Dragan, and S.~Levine, ``Shared autonomy via deep reinforcement
  learning,'' in \emph{Robotics: Science and Systems}, 2018.

\bibitem{javdani2018shared}
S.~Javdani, H.~Admoni, S.~Pellegrinelli, S.~S. Srinivasa, and J.~A. Bagnell,
  ``Shared autonomy via hindsight optimization for teleoperation and teaming,''
  \emph{The International Journal of Robotics Research}, vol.~37, no.~7, pp.
  717--742, 2018.

\bibitem{jain2019probabilistic}
S.~Jain and B.~Argall, ``Probabilistic human intent recognition for shared
  autonomy in assistive robotics,'' \emph{ACM Transactions on Human-Robot
  Interaction}, vol.~9, no.~1, pp. 1--23, 2019.

\bibitem{jonnavittula2021learning}
A.~Jonnavittula and D.~P. Losey, ``Learning to share autonomy across repeated
  interaction,'' in \emph{IEEE/RSJ International Conference on Intelligent
  Robots and Systems}, 2021, pp. 1851--1858.

\bibitem{lynch2020learning}
C.~Lynch, M.~Khansari, T.~Xiao, V.~Kumar, J.~Tompson, S.~Levine, and
  P.~Sermanet, ``Learning latent plans from play,'' in \emph{Conference on
  Robot Learning}, 2020, pp. 1113--1132.

\bibitem{pertsch2020accelerating}
K.~Pertsch, Y.~Lee, and J.~J. Lim, ``Accelerating reinforcement learning with
  learned skill priors,'' in \emph{Conference on Robot Learning}, 2020.

\bibitem{singh2020parrot}
A.~Singh, H.~Liu, G.~Zhou, A.~Yu, N.~Rhinehart, and S.~Levine, ``{PARROT:
  D}ata-driven behavioral priors for reinforcement learning,'' in
  \emph{International Conference on Learning Representations}, 2020.

\bibitem{shankar2020learning}
T.~Shankar and A.~Gupta, ``Learning robot skills with temporal variational
  inference,'' in \emph{International Conference on Machine Learning}, 2020,
  pp. 8624--8633.

\bibitem{merel2018neural}
J.~Merel, L.~Hasenclever, A.~Galashov, A.~Ahuja, V.~Pham, G.~Wayne, Y.~W. Teh,
  and N.~Heess, ``Neural probabilistic motor primitives for humanoid control,''
  in \emph{International Conference on Learning Representations}, 2018.

\bibitem{hausman2018learning}
K.~Hausman, J.~T. Springenberg, Z.~Wang, N.~Heess, and M.~Riedmiller,
  ``Learning an embedding space for transferable robot skills,'' in
  \emph{International Conference on Learning Representations}, 2018.

\bibitem{ajay2020opal}
A.~Ajay, A.~Kumar, P.~Agrawal, S.~Levine, and O.~Nachum, ``{OPAL: O}ffline
  primitive discovery for accelerating offline reinforcement learning,'' in
  \emph{International Conference on Learning Representations}, 2020.

\bibitem{pathak2017curiosity}
D.~Pathak, P.~Agrawal, A.~A. Efros, and T.~Darrell, ``Curiosity-driven
  exploration by self-supervised prediction,'' in \emph{International
  Conference on Machine Learning}, 2017, pp. 2778--2787.

\bibitem{eysenbach2019diversity}
B.~Eysenbach, A.~Gupta, J.~Ibarz, and S.~Levine, ``Diversity is all you need:
  {L}earning skills without a reward function,'' in \emph{International
  Conference on Learning Representations}, 2019.

\bibitem{sharma2020dynamics}
A.~Sharma, S.~Gu, S.~Levine, V.~Kumar, and K.~Hausman, ``Dynamics-aware
  unsupervised discovery of skills,'' in \emph{International Conference on
  Learning Representations}, 2020.

\bibitem{hazan2019provably}
E.~Hazan, S.~Kakade, K.~Singh, and A.~Van~Soest, ``Provably efficient maximum
  entropy exploration,'' in \emph{International Conference on Machine
  Learning}, 2019, pp. 2681--2691.

\bibitem{liu2021behavior}
H.~Liu and P.~Abbeel, ``Behavior from the void: {U}nsupervised active
  pre-training,'' in \emph{Advances in Neural Information Processing Systems},
  vol.~34, 2021.

\bibitem{lee2021pebble}
K.~Lee, L.~Smith, and P.~Abbeel, ``{PEBBLE: F}eedback-efficient interactive
  reinforcement learning via relabeling experience and unsupervised
  pre-training,'' in \emph{International Conference on Machine Learning}, 2021,
  pp. 6152--6163.

\bibitem{muelling2015autonomy}
K.~Muelling, A.~Venkatraman, J.-S. Valois, J.~Downey, J.~Weiss, S.~Javdani,
  M.~Hebert, A.~B. Schwartz, J.~L. Collinger, and J.~A. Bagnell, ``Autonomy
  infused teleoperation with application to {BCI} manipulation,'' in
  \emph{Robotics: Science and Systems}, 2015.

\bibitem{newman2018harmonic}
B.~A. Newman, R.~M. Aronson, S.~S. Srinivasa, K.~Kitani, and H.~Admoni,
  ``Harmonic: {A} multimodal dataset of assistive human-robot collaboration,''
  \emph{arXiv preprint arXiv:1807.11154}, 2018.

\bibitem{haarnoja2018soft}
T.~Haarnoja, A.~Zhou, P.~Abbeel, and S.~Levine, ``Soft actor-critic:
  {O}ff-policy maximum entropy deep reinforcement learning with a stochastic
  actor,'' in \emph{International Conference on Machine Learning}, 2018, pp.
  1861--1870.

\bibitem{doersch2016tutorial}
C.~Doersch, ``Tutorial on variational autoencoders,'' \emph{arXiv preprint
  arXiv:1606.05908}, 2016.

\bibitem{shen2020igibson}
B.~Shen, F.~Xia, C.~Li, R.~Mart{\'\i}n-Mart{\'\i}n, L.~Fan, G.~Wang,
  C.~P{\'e}rez-D’Arpino, S.~Buch, S.~Srivastava, and L.~Tchapmi, ``{iGibson}
  1.0: {A} simulation environment for interactive tasks in large realistic
  scenes,'' in \emph{IEEE/RSJ International Conference on Intelligent Robots
  and Systems}, 2020, pp. 7520--7527.

\bibitem{redmon2016you}
J.~Redmon, S.~Divvala, R.~Girshick, and A.~Farhadi, ``You only look once:
  {U}nified, real-time object detection,'' in \emph{IEEE Conference on Computer
  Vision and Pattern Recognition}, 2016, pp. 779--788.

\end{thebibliography}

\end{document}